\documentclass[letterpaper]{article} 
\usepackage[]{aaai2026}  
\usepackage{times}  
\usepackage{helvet}  
\usepackage{courier}  
\usepackage[hyphens]{url}  
\usepackage{graphicx} 
\usepackage{natbib}  
\usepackage{caption} 
\usepackage{algorithm}
\usepackage{algorithmic}
\usepackage{amsmath}
\usepackage{amssymb}
\usepackage{multirow}
\usepackage{colortbl}
\usepackage{multirow}
\usepackage{tikz}

\usepackage{newfloat}
\usepackage{listings}
\DeclareCaptionStyle{ruled}{labelfont=normalfont,labelsep=colon,strut=off} 
\floatstyle{ruled}
\newfloat{listing}{tb}{lst}{}
\floatname{listing}{Listing}
\title{Dynamic Embedding of Hierarchical Visual Features for Efficient Vision-Language Fine-Tuning}
\author{
    Xinyu Wei\textsuperscript{\rm 1},
    Guoli Yang\textsuperscript{\rm 2}\thanks{Corresponding author.},
    Jialu Zhou\textsuperscript{\rm 1},
    Mingyue Yang\textsuperscript{\rm 1},
    Leqian Li\textsuperscript{\rm 1},
    Kedi Zhang\textsuperscript{\rm 1}
    Chunping Qiu\textsuperscript{\rm 3}
}
\affiliations{
    \textsuperscript{\rm 1}College of Computer Science and Technology, National University of Defense Technology, china\\
    \textsuperscript{\rm 2}Advanced Institute of Big Data, Beijing, China\\
    \textsuperscript{\rm 3}Intelligent Game and Decision Lab, Beijing 100091, China
    
    \{xinyuwei,zhoujialu23,yangmingyue216,llq,zhangkedi10\}@nudt.edu.cn, yanggl@aibd.ac.cn, chunping.qiu@aliyun.com
}

\usepackage{bibentry}

\begin{document}

\maketitle

\begin{abstract}
Large Vision-Language Models (LVLMs) commonly follow a paradigm that projects visual features and then concatenates them with text tokens to form a unified sequence input for Large Language Models (LLMs). However, this paradigm leads to a significant increase in the length of the input sequence, resulting in substantial computational overhead. Existing methods attempt to fuse visual information into the intermediate layers of LLMs, which alleviate the sequence length issue but often neglect the hierarchical semantic representations within the model and the fine-grained visual information available in the shallower visual encoding layers. To address this limitation, we propose DEHVF, an efficient vision-language fine-tuning method based on dynamic embedding and fusion of hierarchical visual features. Its core lies in leveraging the inherent hierarchical representation characteristics of visual encoders and language models. Through a lightweight hierarchical visual fuser, it dynamically selects and fuses hierarchical features corresponding to semantic granularity based on the internal representations of each layer in LLMs. The fused layer-related visual features are then projected and aligned before being directly embedded into the Feed-Forward Network (FFN) of the corresponding layer in LLMs. This approach not only avoids sequence expansion but also dynamically fuses multi-layer visual information. By fine-tuning only a small number of parameters, DEHVF achieves precise alignment and complementarity of cross-modal information at the same semantic granularity. We conducted experiments across various VL benchmarks, including visual question answering on ScienceQA and image captioning on COCO Captions. The results demonstrate that DEHVF achieves higher accuracy than existing parameter-efficient fine-tuning (PEFT) baselines while maintaining efficient training and inference.
\end{abstract}

\section{Introduction}

In recent years, Large Vision-Language Models (LVLMs) have achieved significant success in multimodal understanding tasks by integrating pre-trained visual encoders with Large Language Models (LLMs) \cite{liu2023visual,chen2022pali}. To perceive visual information, the typical approach involves using a projector to align image features with the text semantic space before feeding them into LLMs for inference. However, the computational cost of large models is mainly concentrated in the attention mechanisms of LLMs, whose complexity scales to $\mathcal{O}(n^2)$ with input sequence length $n$. Such methods require concatenating high-dimensional visual feature sequences with text inputs, resulting in sequence length expansion and consequently significant computational overhead.

To address this issue, another approach to building LVLMs involves integrating visual information into the intermediate layers of LLMs. Specifically, two approaches are included: one involves inserting cross-attention layers into LLMs to integrate visual information \cite{alayrac2022flamingo,li2025otter}, and the other uses a designed visual weight generator to align visual features with the text space and directly fuse them with the language model weights \cite{ma2024visual,jie2024memory}, preserving visual information while reducing computational burden. Although effective, we observe two key limitations: 1) Each layer of the language model uses dedicated cross-attention modules, which introduces a large number of new parameters. 2) Different layers of LLMs handle semantic representations at varying granularities—shallow layers focus on local details, while deep layers concentrate on global semantics. Recent research on visual representation learning \cite{walmer2023teaching,ghiasi2022vision} has also proposed the analogous hierarchies in CLIP-ViT, where shallow layers capture basic textures, intermediate layers represent more complex patterns and object parts, and the deepest layers encode complete objects. However, current methods rely solely on the high-level features from the last layer of the visual encoder, injecting aligned features with the same semantic meaning into all layers of the language model. This not only ignores the fine-grained visual information captured by shallow layers (such as edges and textures) but also lacks adaptability to the language model's differentiated hierarchical representations.

\begin{figure*}[]      
	\centering
    \includegraphics[scale=0.77]{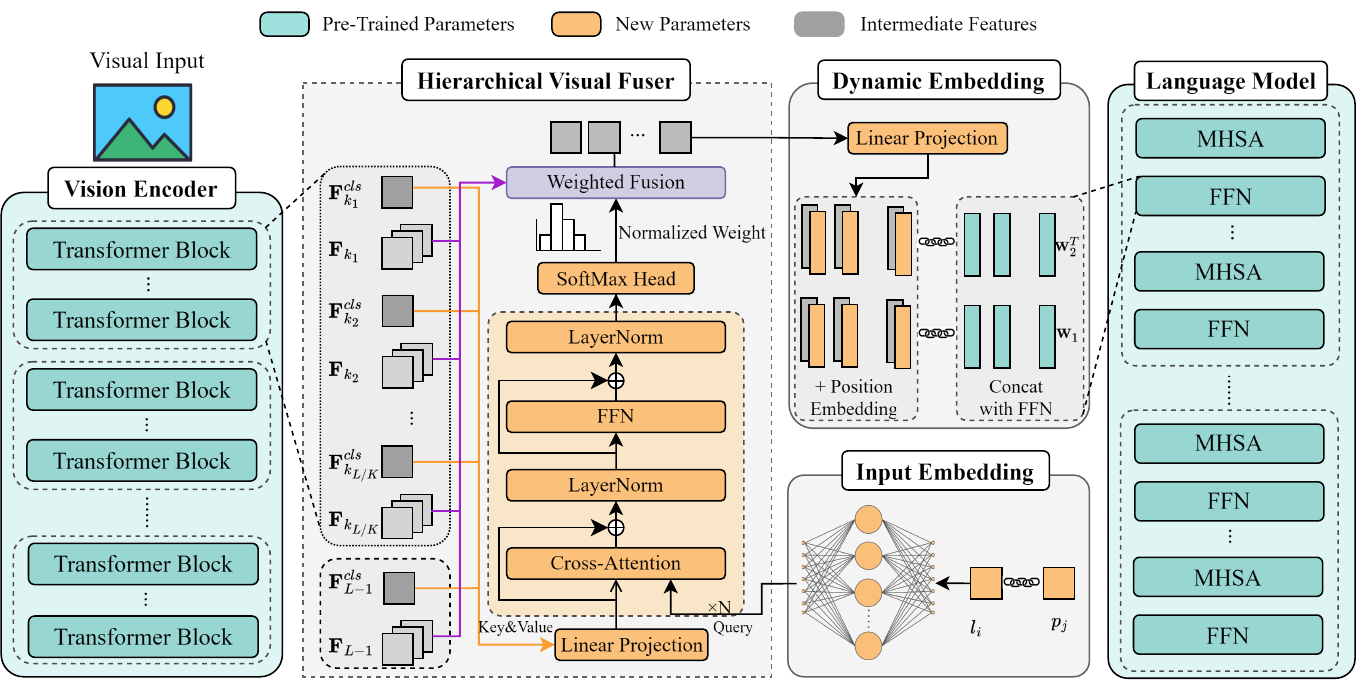}
	\caption{Overall Framework. Composed of five main modules: visual encoder, input embedding network, hierarchical visual fuser, dynamic embedding, and LLMs.}
	\label{framework}
\end{figure*}

Neural network representation learning generally exhibits hierarchical characteristics. We argue that the features of different layers in language models and their corresponding layers in visual encoders have inherent compatibility in terms of semantic granularity and representational hierarchy. Therefore, we consider injecting features from specific layers of the visual encoder into the corresponding layers of the language model to achieve more effective and flexible cross-modal information fusion. Based on this, we propose a parameter-efficient visual-language fine-tuning method, namely the Dynamic Embedding of Hierarchical Visual Features for Efficient Vision-Language Fine-Tuning (DEHVF). As shown in Figure \ref{framework}, we first input the hierarchical representations of the language model obtained through the input embedding network into the hierarchical visual fuser, dynamically selecting and integrating the fused visual features corresponding to the semantic granularity for each language model layer. Subsequently, these layer-related weighted visual features are aligned with the text space via a projector and fused with position embeddings, which are directly concatenated with the FFN weight matrices in corresponding language model layers. During fine-tuning, we freeze most parameters of the visual encoder and language model, only the input embedding network, VL projection layer, hierarchical visual fuser, and position embeddings are tunable. The DEHVF not only avoids increasing the input sequence length but also dynamically fuses multi-layer visual information, effectively adapting to the natural hierarchical structure of each modality's internal representations. This enables precise alignment and complementarity of cross-modal information at the same semantic granularity, improving model performance while maintaining efficient training and inference.

Our contributions are summarized as follows:

\begin{itemize}
\item We designed a lightweight hierarchical representation-guided hierarchical visual fuser. By establishing a visual-language hierarchical mapping, we dynamically assign weights to visual features based on the hierarchical representation of the language model, achieving adaptive fusion of visual features within different hierarchical levels. This enables the selection of visual features that better match the hierarchical semantics of the language model.
\item We propose an embedding mechanism that fully leverages the hierarchical characteristics of visual encoders and language models, dynamically concatenating hierarchical visual features into the weight matrix of LLMs. This not only avoids increasing the input sequence length but also enables dynamic fusion of multi-layer visual information, thereby achieving precise alignment and complementarity of cross-modal information at the same semantic granularity, effectively improving model accuracy and inference speed.
\item We conducted comprehensive experiments on the publicly available ScienceQA and COCO Caption datasets. The experimental results demonstrate the effectiveness and efficiency of DEHVF.
\end{itemize}

\section{Related Works}
\subsection{Multimodal architectures}

Recent advances in deep learning have significantly propelled the flourishing development of multimodal intelligence \cite{baltruvsaitis2018multimodal,li2023variational,zhang2022magic,xiao2024pointblip}. LVLMs leverage the powerful capabilities of LLMs to avoid additional computational overhead, thereby significantly enhancing the efficiency of multimodal pretraining \cite{zhang2024mm,lan2024efficient}.The visual encoder, typically based on CLIP-Vision Transformers (CLIP-ViTs), extracts visual features from input images, which are then projected into the input embedding space of the LLMs by the vision-language projector. Finally, the LLM jointly processes the aligned visual features with text instructions to generate the final response. LVLMs can be broadly categorized into two classes based on how visual information is integrated into the pre-trained LLMs.

One class of models introduce visual information at the input stage of the language model. In this approach, the extracted visual information is typically concatenated with language tokens before being input into the LLMs. Typically, a projection layer is used to align the extracted visual feature space with the language token space, where the projection layer can be a simple linear layer \cite{liu2024improved,driess2023palm,chen2023minigpt}, a resampling layer \cite{li2023blip}\cite{zhu2023minigpt}, etc.

To address efficiency issues caused by the increased length of the language model input sequence, another class of models integrate visual information into the inner layers of the language model. In this category, a pre-trained visual encoder is first used to extract visual information, which is then processed through learnable projection layers or cross-attention mechanisms. Multimodal fusion is achieved by inserting cross-attention layers into the LLMs, \cite{alayrac2022flamingo,li2025otter} directly merging with LLM weights \cite{ma2024visual}, or concatenating with the weight matrices of FFN layers in Transformer modules \cite{jie2024memory}.

\subsection{Parameter-efficient fine-tuning}

As the parameter scale of large language models continues to grow, Parameter-Efficient Fine-Tuning (PEFT) alleviates the enormous training and storage burdens associated with full fine-tuning. According to whether new trainable parameters are introduced into pretrained models, PEFT for language models can be broadly categorized into two types, additive fine-tuning methods and Low-Rank Adaption (LoRA). Additive fine-tuning involves adding extra modules or parameters (such as adapter \cite{houlsby2019parameter,lei2023conditional,chronopoulou2023adaptersoup}  and prefix-tuning \cite{li2021prefix,liu2021p,liu2024gpt,zhu2023spt}), training only the newly added parameters while freezing the pre-trained model. LoRA methods construct low-rank weight matrices and merge with the original weights, updating only these low-rank matrices.

In addition to the general PEFT schemes, there are many PEFT methods specifically designed for multimodal learning. VL-Adapter \cite{sung2022vl} and VL-PET\cite{hu2023vl} are based on an encoder-decoder architecture, achieving efficient cross-modal parameter transfer by inputting visual features into the input layer and fine-tuning an adapter module; LLaMA-Adapter \cite{zhang2023llama} and LaVIN \cite{luo2023cheap} enable multimodal interaction by designing lightweight adapters for the LLaMA \cite{touvron2023llama}; UniAdapter \cite{lu2023uniadapter} proposes a knowledge-sharing adapter design that effectively adapts to cross-modal representations. Aurora \cite{wang2023parameter} employs modal approximation to achieve a prompt framework for efficient multimodal transfer learning. Wander \cite{guo2025wander} enables fine-grained token-level interaction between sequences of different modalities. MemVP \cite{jie2024memory} innovatively achieves multimodal fusion by injecting visual features into the FFN of language models.

\section{Method}
The overall framework of our method is illustrated in Figure \ref{framework}. It consists of five main modules: a visual encoder, an input embedding network, a hierarchical representation-guided hierarchical visual fuser (HVF), a dynamic embedding, and a large language model (LLM).

\subsection{Overall Framework}

\textbf{Vision Encoder.} We adopt CLIP-ViT \cite{radford2021learning} as the image encoder. The input image $\mathbf{I}\in\mathbb{R}^{C\times H\times W}$ is partitioned into $N$ fixed-size patches. After linear projection, these are concatenated with a learnable $<cls>$ token to form the initial embeddings $\mathbf{Z} \in \mathbb{R}^{(N+1) \times D}$. This sequence is processed by $L_v$ Transformer layers to extract visual features. Each layer of the visual encoder produces $N+1$ representations. The hierarchical output of the visual encoder is:
\begin{equation}  
\mathbf{F}=Transformer^{(L_v)}(\mathbf{Z})\in\mathbb{R}^{L_v\times(N+1)\times D}
\end{equation}
where $L_v$ is the number of transformer layers, $N$ is the number of patches, and $D$ is the hidden state dimension. Each layer produces $N+1$ representations, because CLIP-ViT appends a special $<cls>$ token to the patch embeddings to obtain a global representation.

\textbf{Input Embedding Network.} We first obtain the hierarchical representation $I_{l,p}\in\mathbb{R}^t$ for each layer of language models using an input embedding network $h(.)$, which is a multilayer perceptron (MLP) composed of two feedforward layers and an activation function. For a Transformer model with $L_t$ layers, we introduce a set of learnable layer id embeddings $\mathcal{L}=\{l_{i}\}_{i=1}^{L_t}$ and position embeddings $\mathcal{P}=\{p_{j}\}_{j=1}^{2}$ that distinguish the upload and download projections of the FFN. The concatenation of $l_i\in\mathbb{R}^t$ and $p_j\in\mathbb{R}^t$, denoted as $(l_i,p_{j})$, is the input to the network. The input embedding network $h(.)$ learns appropriate compressed embedding features from the input:
\begin{equation}  
I_{l,p}=h({l_i},{p_{j}})
\end{equation}
The obtained compressed embedding features, serving as representations for specific layers and positions, are fed into the Hierarchical Visual Fuser to guide subsequent visual feature fusion.

\textbf{Hierarchical Visual Fuser.} Each layer in LLMs focuses on different visual information, so we design a hierarchical visual fuser that dynamically integrates visual features based on layer-specific input embeddings. To align with the inherent hierarchical representations of both the visual encoder and the language model, this fuser first needs to establish a visual-language hierarchy mapping for selecting cross-modal features of corresponding granularity. Based on recent research \cite{li2025instruction} analyzing the performance of hierarchical visual features, we divide the $L_v$ Transformer layers of the visual encoder into $K$ visual groups. Each group captures visual information of a specific granularity  ($k=1$ corresponds to the shallowest layer/fine-grained information; $k=K$ corresponds to the deepest layer/semantic information), with each visual group containing $L_v/K$ consecutive layers. To maintain semantic richness, each visual group adds features from the penultimate layer, forming the final visual groupings, each of which includes $L_v/K + 1$ hierarchical visual features:

\begin{equation}  
\mathbf{F}_{k}^{cls}=\{\mathbf{F}^{cls}_{k_{1}},...,\mathbf{F}^{cls}_{k_{L/K}},\mathbf{F}^{cls}_{L-1}\},\quad k=1,2,...,K
\end{equation}
\begin{equation}  
\mathbf{F}_{k}=\{\mathbf{F}_{k_{1}},...,\mathbf{F}_{k_{L/K}},\mathbf{F}_{L-1}\},\quad k=1,2,...,K
\end{equation}
where $\mathrm{cls}_{k_i}\in\mathbb{R}^D$ is the $<cls>$ feature of the $i$-th layer in group $k$, and $\mathbf{F}_{k_{i}}\in\mathbb{R}^{N\times D}$ is the visual feature of the $i$-th layer in group $k$.

Similarly, we divide the language model's $L_t$ Transformer layers into $K$ groups corresponding to the visual groups, with each group containing $L_t/K$ consecutive layers. Thus, the group corresponding to layer $l$ is $k =\left \lfloor l/ (L_{t}/K ) \right \rfloor$. In this way, for layers $l=1, 2, \dots, L_t$ of the language model, we use this layering mechanism to dynamically select visual information of the corresponding granularity at different abstraction levels of language processing.

Additionally, the hierarchical visual fuser needs to fuse $L_v/K + 1$ hierarchical visual features of group $k$ associated with layer $l$. To generate adaptive fusion features, we dynamically compute fusion weights for visual group $k$ based on layer-specific embeddings. As shown in Figure \ref{framework}, the hierarchical visual fuser consists of a linear projection layer, $N$ Transformer blocks, and a Softmax network. Each Transformer block contains a cross-attention module and a feed-forward network, with outputs of each module processed by residual connections and layer normalization. Specifically, grouped $<cls>$ features are first mapped to the fusion layer's hidden dimension through linear projection. Then the compressed embedding feature $I_{l,p}\in\mathbb{R}^t$ generated by the input embedding network is used as layer instruction Query, and the projected hierarchical global representations $\mathbf{F}_{k}^{cls}$ are used as the Key/Value, and calculate the importance of different layer visual features $\mathbf{F}_{k_{1}},\mathbf{F}_{k_{2}}, \dots ,\mathbf{F}_{k_{L/K}},\mathbf{F}_{L-1}$ in visual group $\mathbf{F}_{k}$ corresponding to layer $l$ through cascaded Transformer blocks. Finally, the Softmax network generates the normalized weight vector $\mathbf{w}^{l,p}\in\mathbb{R}^{L/K+1}$:

\begin{equation} 
\begin{aligned}
\mathbf{w}^{l,p}&= HVF(I_{l,p},\mathbf{F}_{k}^{cls}) \\
&=HVF(I_{l,p},\mathbf{F}^{cls}_{k_{1}},\mathbf{F}^{cls}_{k_{2}},...,\mathbf{F}^{cls}_{k_{L/K}},\mathbf{F}^{cls}_{L-1})
\end{aligned}
\end{equation}

where $\sum_{i=1}^{L/K+1}w_i^{l,p}=1$, $k=\left \lfloor l/(L_{t}/K) \right \rfloor$.

Based on this weight vector, the hierarchical fused feature is obtained by weighted summation of the visual features of each layer within group $k$:
\begin{equation}  
\mathbf{F}^{l,p}=\sum_{i=1}^{L/K+1}w_i^{l,p}\times\mathbf{F}_{k_i},\quad l=1,...,L_t;p=0,1
\end{equation}

\textbf{Dynamic Embeddings.} The standard FFN in the language model consists of two fully connected layers. Let $x\in\mathbb{R}^{h}$ be the input token to the FFN, and $\phi$ be the activation function. The FFN can then be expressed as:
\begin{equation}  
\mathrm{FFN}(\boldsymbol{x})=\phi(\boldsymbol{x}\boldsymbol{W}_1)\boldsymbol{W}_2
\end{equation}

Here, $W_1\in\mathbb{R}^{h\times H}$ and $W_{2}\in\mathbb{R}^{H\times h}$ are weight matrices of two fully connected layers. Based on the fact that the FFN in pretrained language models essentially functions as a key-value memory storing factual knowledge\cite{geva2020transformer}, "basketball is spherical" might be such knowledge stored in FFN. Thus $W_1$ and $W_{2}$ can be deconstructed into key-value pairs:

\begin{equation}  
\boldsymbol{W}_1=(\boldsymbol{k}_1,\boldsymbol{k}_2,...,\boldsymbol{k}_H),\boldsymbol{W}_2=(\boldsymbol{v}_1,\boldsymbol{v}_2,...,\boldsymbol{v}_H)^\intercal
\end{equation}

The FFN is equivalent to a retrieval activation weighting process of stored knowledge in key-value memory:
\begin{equation}  
\mathrm{FFN}(\boldsymbol{x})=\sum_{i=1}^D\phi(\langle\boldsymbol{x},\boldsymbol{k}_i\rangle)\cdot\boldsymbol{v}_i
\end{equation}

In language models, FFN only stores text knowledge acquired during pretraining. Inspired by MemVP\cite{jie2024memory}, we propose a hierarchy-aware dynamic embedding mechanism to inject visual features into FFN, supplementing vision-related knowledge previously lacking in language models, e.g., enabling the model to recognize "objects in images are spherical". Visual features should align with the dimension of text tokens while preserving positional information. Therefore, the fused visual feature $\mathbf{F}^{l,p}=(f_1^{l,p},f_2^{l,p},...,f_n^{l,p})^\intercal\in\mathbb{R}^{N\times D}$ from the hierarchical visual fuser is projected to text token dimensions via projector $g$ (which may consist of one or more fully connected layers). The projected visual features are then summed with positional embeddings:
\begin{equation}  
\mathcal{K}(\mathbf{F}^{l,p})=\lambda g(\mathbf{F}^{l,p})+\boldsymbol{s}^k,\quad p=0
\end{equation}
\begin{equation}  
\mathcal{V}(\mathbf{F}^{l,p})=\lambda g(\mathbf{F}^{l,p})+\boldsymbol{s}^v,\quad p=1
\end{equation}

where $\lambda$ is a hyperparameter, and $s^{k},s^{v}\in\mathbb{R}^{N\times D}$ are the visual prompt positional embeddings inserted into the key and value. When $p=0$ and $p=1$, $\mathbf{F}^{l,p}$ denotes the visual features inserted into the key and value, respectively.

Specifically, to achieve alignment between visual features and language model weights at the same semantic granularity, our $\mathbf{F}^{l,p}$ is hierarchy-aligned visual feature for language model layers, obtained from the hierarchical visual fuser, and then injected into different layers of the FFN:
\begin{equation}  
\mathrm{FFN}^{l}(\boldsymbol{x})=\sum_{i=1}^H\phi(\langle\boldsymbol{x},\boldsymbol{k}_i\rangle)\cdot\boldsymbol{v}_i+\sum_{i=1}^N\phi(\langle\boldsymbol{x},\mathcal{K}(f_i^{l,0})\rangle)\cdot\mathcal{V}(f_i^{l,1})
\end{equation}

Thus for the $l$-th layer FFN, the weight matrix is modified as:
\begin{equation}  
\begin{aligned}&\boldsymbol{{W'}}_{1}=(\boldsymbol{k}_{1},\boldsymbol{k}_{2},...,\boldsymbol{k}_{H},\lambda g(\boldsymbol{f}_{1}^{{l,0}})+\boldsymbol{p}_{1}^{k},...,\lambda g(\boldsymbol{f}_{n}^{{l,0}})+\boldsymbol{p}_{n}^{k}),\\&\boldsymbol{{W'}}_{2}=(\boldsymbol{v}_{1},\boldsymbol{v}_{2},...,\boldsymbol{v}_{H},\lambda g(\boldsymbol{f}_{1}^{{l,1}})+\boldsymbol{p}_{1}^{v},...,\lambda g(\boldsymbol{f}_{n}^{{l,1}})+\boldsymbol{p}_{n}^{v})^{\intercal}.\end{aligned}
\end{equation}

\textbf{Large Language Models.} Hierarchical visual features are projected to the language model space, achieving cross-modal fusion during the FFN forward computation process, and ultimately generating text responses to complete end-to-end visual-language reasoning tasks.

\begin{figure*}      
	\centering
    \includegraphics[scale=0.39]{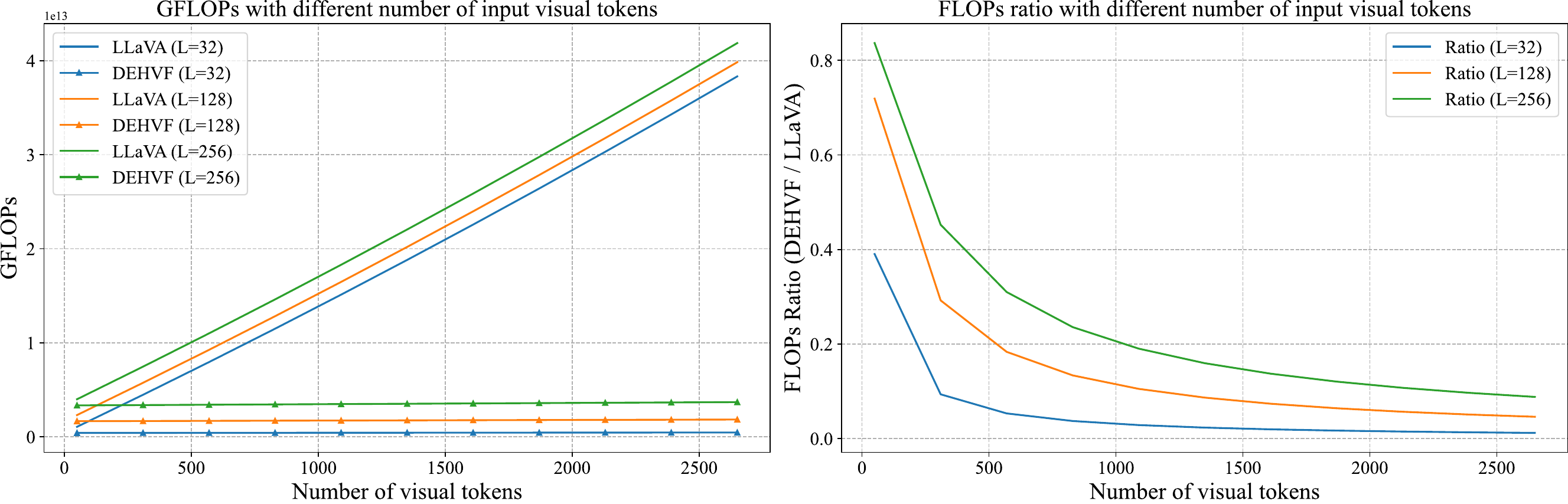 } 
	\caption{Comparison of FLOPs. This figure shows the FLOPs of LLaVA and DEHVF with different numbers of input visual tokens. The left subplot illustrates the change in GFLOPs, the right subplot plots the ratio of GFLOPs for DEHVF to LLaVA, and C denotes the number of text tokens.} 
    \label{FLOPS}
\end{figure*}

\subsection{Computational Cost Analysis}
We only consider the computational cost of the self-attention module and feedforward network in LLMs, omitting the bias and normalization layers. Let $d$, $h$, $L$, and $n$ represent the number of Transformers in LLMs, the hidden state dimension, the text input length, and the visual prompt length, respectively. The FLOPs for the multi-head attention mechanism and the feedforward network (FFN) are $8Lh^2+4L^2h$ and $16Lh^2$. For LLMs that concatenate visual prompts to text inputs, the input sequence length becomes $L+n$, with FLOPs of $24d(L + n)h^2 + 4d(L + n)^2h$. The additional computational cost of DEHVF primarily stems from three components: the input embedding network, the hierarchical visual fuser, and the increased hidden dimension of the FFN. Assuming the hierarchical visual fuser in DEHVF contains k Transformer Blocks with a hidden layer dimension of $r$, and based on our configuration, the input embedding network has input, hidden, and output dimensions of $2r$, $2r$, and $r$, respectively. According to calculations, the FLOPs of the input embedding network are $12r^3$, the FLOPs in the hierarchical visual fusion module are approximately $3r^3 + 8kr^3$, and the additional FLOPs from the visual features concatenated to the FFN are $4dLnh$.

Detailed FLOPs calculations are provided in Appendix A. In Figure \ref{FLOPS}, we compare the FLOPs of LLaVA and our method. As the number of visual tokens increases, DEHVF does not introduce additional quadratic computation. When the text length is 32, our FLOPs are only 6\% of LLaVA-v1.5.

\section{Experiments and Results}
Prior studies demonstrate that even without resource-intensive vision-language pretraining, the combination of a vision encoder and language model can still be transferred to downstream vision-language tasks via PEFT while matching the performance of full fine-tuning. Given the practical need for low-cost deployment, we build on prior work \cite{luo2023cheap} and omit vision-language pretraining in experiments, employing an efficient lightweight adaptation setup.

\subsection{Experimental Setup}
\textbf{Datasets.} We evaluate model performance using the ScienceQA dataset \cite{lu2022learn}. This is a large-scale multimodal science QA dataset originates from elementary and middle school science curricula, covering 26 subtopics and 379 skills across natural science, language science, and social science domains. Divided into train/val/test splits with 12,726/4,241/4,241 examples respectively. We use average accuracy for evaluation. In addition, we use the Karpathy splits \cite{karpathy2015deep} of COCO Caption dataset \cite{chen2015microsoft} to evaluate DEHVF, using BLEU-4 and CIDEr metrics to quantify the semantic alignment between the generated text and the manually annotated text.

\textbf{Models.} We use LLaMA \cite{touvron2023llama} with 7B and 13B parameters as the language model. The visual encoder is CLIP-ViT-L/14 \cite{radford2021learning,dosovitskiy2020image}.

\textbf{Implementation Details.} We follow LaVIN's \cite{luo2023cheap} optimization configuration. Specifically, we use AdamW \cite{loshchilov2017decoupled} optimizer with cosine decay learning rate schedule, training for 20 epochs on ScienceQA and 5 epochs on COCOCaption. Batch size, learning rate, and weight decay are set to 4, 2e-2, and 0.02 respectively. We compare DEHVF with baseline method MemVP and other LLaMA-based fine-tuning approaches including LLaVA, LLaMA-Adapter, and LaVIN. We also include LLaVA fine-tuned with LoRA.

For projector design, DEHVF, MemVP, LaVIN, and LLaVA-LoRA all employ a two fully connected layers with intermediate nonlinear activation. Since LaVIN introduces additional adapters in the visual encoder, for fair comparison we integrate similar modules in MemVP, LLaVA-LoRA, and DEHVF. Specifically, we add parallel adapters to the visual encoder's FFN. The hierarchical visual fuser contains 4 cross-attention Transformers, each with a hidden dimension of 64 and 4 attention heads. We partition the visual encoder's 24 layers and language model's 32 layers into 4 distinct visual groups.

\begin{table*}[]
\small
\centering

\begin{tabular}{ccc|ccccccccc}
\hline
\multirow{2}{*}{Method} & \multirow{2}{*}{Param} & \multirow{2}{*}{\begin{tabular}[c]{@{}c@{}}Laguage\\ Model\end{tabular}} & \multicolumn{3}{c}{Subject}                      & \multicolumn{3}{c}{Context Modality}             & \multicolumn{2}{c}{Grade}       & \multirow{2}{*}{avg} \\
                        &                        &                                                                          & NAT            & SOC            & LAN            & TXT            & IMG            & NO             & G1-6           & G7-12          &                      \\ \hline
Human                   & -                      & -                                                                        & 90.23          & 84.97          & 87.48          & 89.60          & 87.50          & 88.10          & 91.59          & 82.42          & 88.40                \\
GPT-4                   & -                      & GPT-4                                                                    & 84.06          & 73.45          & 87.36          & 81.87          & 70.75          & 90.73          & 84.69          & 79.10          & 82.69                \\
LLaVA                   & 7B                     & Vicuna-7B                                                                & -              & -              & -              & -              & -              & -              & -              & -              & 89.84                \\
LLaVA                   & 13B                    & Vicuna-13B                                                               & -              & -              & -              & -              & -              & -              & -              & -              & 85.81                \\
LLaVA                   & 13B                    & Vicuna-13B                                                               & 90.36          & 95.95          & 88.00          & 89.49          & 88.00          & 90.66          & 90.93          & 90.90          & 90.92                \\ \hline
PEFT methods            &                        &                                                                          &                &                &                &                &                &                &                &                &                      \\
LLaMAAdapter            & 1.8M                   & LLaMA-7B                                                                 & 84.73          & 88.30          & 84.36          & 83.72          & 80.32          & 86.90          & 85.83          & 84.05          & 85.19                \\
LLaVA-LoRA              & 4.4M                   & LLaMA-7B                                                                 & 91.70          & 94.60          & 86.09          & 91.25          & 90.28          & 88.64          & 91.52          & 89.65          & 90.85                \\
LaVIN                   & 3.8M                   & LLaMA-7B                                                                 & 89.25          & 94.94          & 85.24          & 88.51          & 87.46          & 88.08          & 90.16          & 88.07          & 89.41                \\
MemVP                   & 3.9M                   & LLaMA-7B                                                                 & 94.45          & 95.05          & 88.64          & 93.99          & 92.36          & 90.94          & 93.10          & 93.01          & 93.07                \\
\rowcolor{gray!25} 
DHVEF(ours)             & 4.2M                   & LLaMA-7B                                                                 & 95.03          & 94.94          & 89.36          & 94.62          & 93.11          & 91.50          & 93.28          & 94.00          & 93.54                \\
LaVIN                   & 5.4M                   & LLaMA-13B                                                                & 90.32          & 94.38          & 87.73          & 89.44          & 87.65          & 90.31          & 91.19          & 89.32          & 90.50                \\
MemVP                   & 5.5M                   & LLaMA-13B                                                                & 95.07          & 95.15          & 90.00          & 94.43          & 92.86          & 92.47          & 93.61          & 94.07          & 93.78                \\
\rowcolor{gray!25} 
DHVEF(ours)             & 5.8M                   & LLaMA-13B                                                                & \textbf{95.74} & \textbf{95.39} & \textbf{90.91} & \textbf{95.11} & \textbf{93.46} & \textbf{92.96} & \textbf{94.27} & \textbf{94.59} & \textbf{94.19}       \\ \hline
\end{tabular}
\caption{ Accuracy on ScienceQA test set. Question categories: NAT = natural science, SOC = social science, LAN = language science, TXT = w/ text context, IMG = w/ image context, NO = no context, G1-6 = grades 1-6, G7-12 = grades 7-12. Except for the results of our method, all other results are quoted from the original papers.}
\label{scienceqa}
\end{table*}

\subsection{Qualitative Experimental Results}

\textbf{ScienceQA.} Table \ref{scienceqa} shows the evaluation results of the DHVEF method on the ScienceQA dataset. The experimental results show that DEHVF outperforms all baseline PEFT methods and the LLaVA model (a full fine tuning model with VL pre-training) on both LLaMA-7B and LLaMA-13B. Among the PEFT methods, DEHVF achieves better performance than the best baseline while maintaining a comparable number of learnable parameters. Specifically, with the LLaMA-7B backbone, our method achieves an average accuracy of 93.54\%, which is 0.47\% higher than the second-best method, with only 0.3M additional parameters. When using LLaMA-13B as the LLM, the method's average accuracy improves to 94.19\%, a 0.41\% gain over the second-best result.

Additionally, we compare the training and inference speeds of different PEFT methods in Table \ref{speed}. We set the batch size to 4 during training and 32 during inference, using NVIDIA/A100 to measure speed. Evaluation results show that although our method is slightly slower than two baseline methods because hierarchical visual fusion at corresponding granularity delays our training and inference, our computational efficiency is significantly higher than traditional PEFT methods like LoRA. DEHVF is 1.5× faster than LLaVA-LoRA during training and inference. Overall, our approach achieves a better balance between computational efficiency and performance.

\begin{table}[]
\small
\centering
\begin{tabular}{cccccc}
\hline
Method        & Params & \begin{tabular}[c]{@{}c@{}}Training \\  Time\end{tabular} & \begin{tabular}[c]{@{}c@{}}Inference \\  Time\end{tabular} & avg   \\ \hline
LLaVA-LoRA 7B & 4.4M   & 0.49                                                      & 2.06                                                       & 90.85 \\
LaVIN 7B      & 3.8M   & 0.39                                                      & 1.24                                                       & 89.41 \\
MemVP 7B      & 3.9M   & 0.28                                                      & 1.13                                                       & 93.07 \\
\rowcolor{gray!25} 
Ours 7B       & 4.2M   & 0.32                                                      & 1.35                                                       & 93.54 \\ \hline
\end{tabular}
\caption{ Training and inference time. Measured using 1×A100 GPU on the ScienceQA dataset, the batch size is 4 for training and 32 for inference.}
\label{speed}
\end{table}

\textbf{COCO Caption.} In Table \ref{cococaption}, we compare DHVEF with existing methods on the image captioning task. From these results, we can still observe the competitive performance of our method. As a PEFT method, our method significantly outperforms LLaMA-Adapterv2\cite{chen2022adaptformer} and LaVIN\cite{luo2023cheap}, with 0.9 and 0.7 improvements in the BLEU-4 score and 6.7 and 2.0 increases in the CIDEr score, respectively. Compared to large-scale pre-trained models such as BLIP\cite{li2022blip} and BLIP-2\cite{li2023blip}, our method's performance remains comparable while being significantly more cost-effective. For example, we only update 4.2M parameters and require only 40 GPU hours on a single A100, whereas BLIP-2 requires over 300 GPU hours on 16 A100s. These results further validate the superiority and efficiency of DHVEF.

\begin{table}[]
\small
\centering
\begin{tabular}{ccccc}
\hline
Method            & PT Data & \#T-Params & BLEU-4 & CIDEr \\ \hline
ClipCap           & 0       & —          & 33.5   & 113.1 \\
BLIP              & 14M     & 583M       & 40.4   & 136.7 \\
BLIP-2            & 129M    & 188M       & 43.7   & 145.3 \\
*Adapter V2       & 0       & 14M        & 36.2   & 122.2 \\
*LaVIN            & 0       & 5.4M       & 36.4   & 126.9 \\
\rowcolor{gray!25} 
ours              & 0       & 5.8M       & 37.1   & 128.9 \\ \hline
\end{tabular}
\caption{Evaluation Results on COCO Caption. *PEFT METHODS.}
\label{cococaption}
\end{table}

\begin{table*}[]
\small
\centering
\begin{tabular}{ccccccccccc}
\hline
\multirow{2}{*}{Settings} & \multirow{2}{*}{\#Trainable} & \multicolumn{3}{c}{Subject} & \multicolumn{3}{c}{Context Modality} & \multicolumn{2}{c}{Grade} & \multirow{2}{*}{average} \\
                          &                              & NAT     & SOC     & LAN     & TXT        & IMG        & NO         & G1-6        & G7-12       &                          \\ \hline
DEHVF 7B                  & 4.17M                        & 95.03   & 94.94   & 89.36   & 94.62      & 93.11      & 91.50      & 93.28       & 94.00       & 93.54                    \\
w/o visual prompts        & 3.22M                        & 86.37   & 78.52   & 88.55   & 85.24      & 76.05      & 91.08      & 85.79       & 84.38       & 85.29                    \\
w/o adaptive weights      & 3.88M                        & 93.21   & 94.94   & 89.64   & 92.77      & 91.62      & 91.57      & 92.51       & 92.88       & 92.64                    \\
w/o multi-features        & 3.88M                        & 94.45   & 95.05   & 88.64   & 93.99      & 92.36      & 90.94      & 93.10       & 93.01       & 93.07                    \\
keys/values same          & 4.15M                        & 92.41   & 94.71   & 88.55   & 91.89      & 89.99      & 91.08      & 92.22       & 91.30       & 91.89                    \\
prompts in keys only      & 4.17M                        & 93.83   & 92.58   & 88.82   & 93.45      & 90.68      & 91.15      & 91.96       & 92.81       & 92.27                    \\
prompts in values only    & 4.17M                        & 93.83   & 95.28   & 89.00   & 93.30      & 91.67      & 91.57      & 92.47       & 93.61       & 92.88                    \\ \hline
\end{tabular}
\caption{Ablation experiments on ScienceQA.}
\label{Ablation}
\end{table*}

\begin{figure}      
	\centering
    \includegraphics[width=1.0\linewidth]{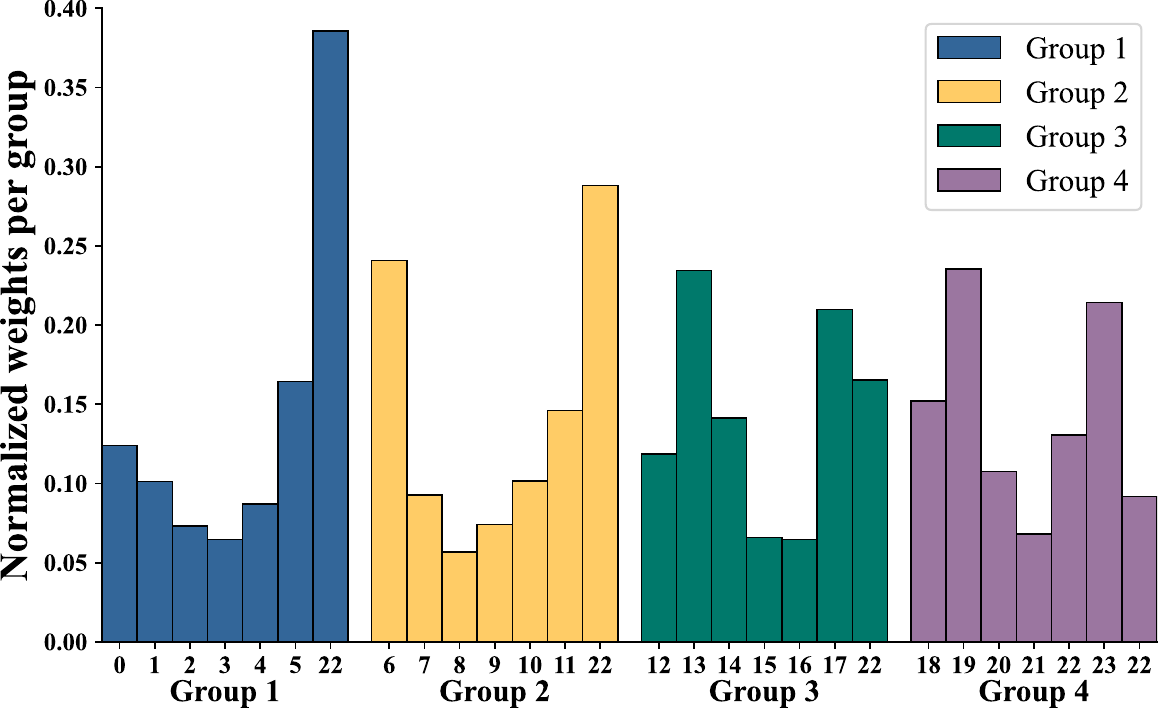} 
	\caption{Average normalized weights per visual group in the ScienceQA test datasets.} 
    \label{weights}
\end{figure}

\begin{figure}      
	\centering
    \includegraphics[width=1.0\linewidth]{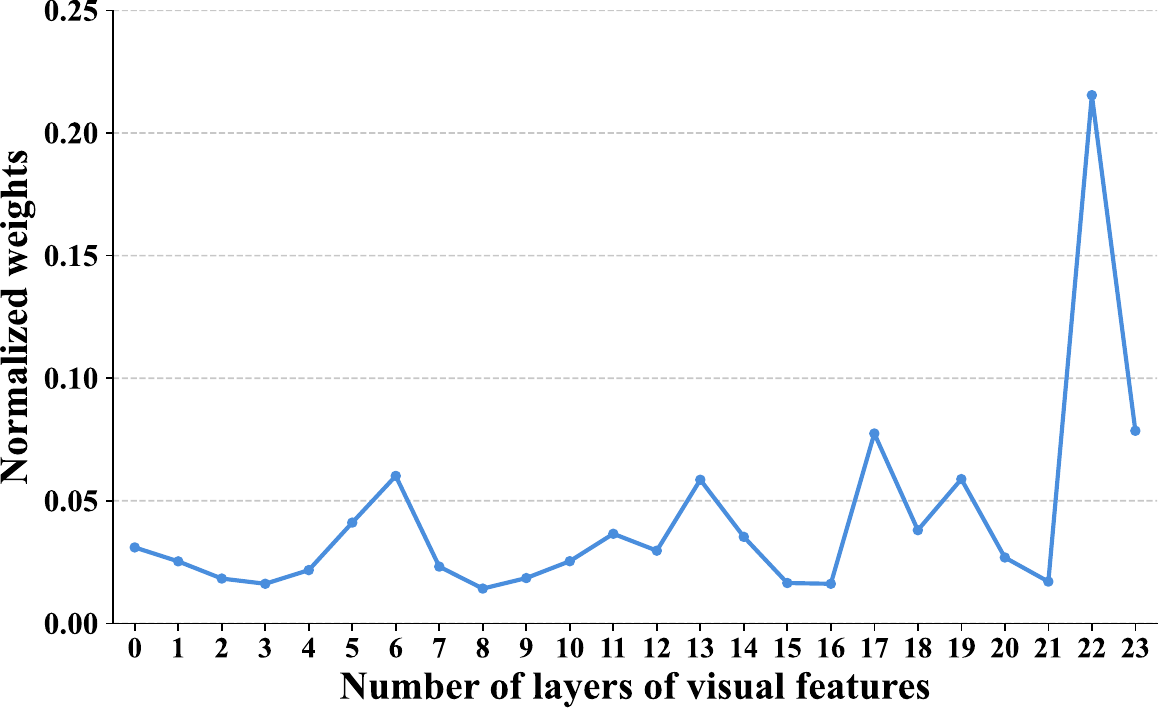 } 
	\caption{Average normalized weights in the ScienceQA test datasets.} 
    \label{line_chart}
\end{figure}

\textbf{Visualization.} In Figure \ref{weights}, we visualize the average normalized weights of visual features across each layer in the visual-language hierarchical mapping groups on the ScienceQA test dataset. For example, for layers 0 to 7 of language model Group 1, we obtain the weights of visual features in visual group 1 corresponding to each layer and then calculate the average. As shown in the figure, the proportion of visual features from the penultimate layer increases in shallower layers, demonstrating the dominant role of higher-level visual features in global reasoning and the rationality of incorporating relevant visual information into the visual group. Additionally, as shown in Figure \ref{line_chart}, we visualized the proportion of visual features at each layer during the language model's reasoning process. Although higher-level visual features account for a higher proportion, the model does not rely on a single higher-level feature but dynamically activates visual information from different layers. The weight distribution curve reveals a multi-peak characteristic, indicating that the model can dynamically select visual features of corresponding granularity based on the language processing stage, validating the effectiveness of the cross-modal same-granularity alignment mechanism.

\subsection{Ablation Study}
To validate the effectiveness of each component in the proposed framework, we conducted a comprehensive ablation study. We designed ablation experiments using the ScienceQA dataset on the LLaMA-7B model to investigate the effectiveness of each module. As shown in Table \ref{Ablation}, when we only insert position embeddings without adding visual prompts to the language model, its performance significantly degrades because the language model cannot access visual knowledge. When we replaced the hierarchical visual fusion mechanism guided by hierarchical representations with a simple average pooling mechanism for visual feature fusion, performance declined, highlighting the importance of dynamically assigning weights to hierarchical features. We also attempted to remove multi-layer feature fusion, causing the model to revert to using only the second-to-last layer features, resulting in performance decline but still achieving good experimental results, further demonstrating the dominant role of high-level visual features in the method. When we replace hierarchical representations with learnable tokens, performance significantly decreases, highlighting the importance of compressed features obtained from the input embedding network. Finally, when visual prompts are inserted into either the key or value, the model's performance decreases in both cases. These results indicate that each component in DHVEF contributes to improving the performance of fine-tuned LVLMs. The combination of all components yields the best results, demonstrating the effectiveness of our design.

\section{Conclusion}
In this paper, we propose a novel dynamic hierarchical visual embedding framework (DHVEF) for efficiently constructing visual language models, based on the natural compatibility between language model features and visual encoder features at corresponding layers in terms of semantic granularity and representation properties. DHVEF dynamically selects and fuses multi-layer features from the visual encoder through a hierarchical visual fuser, and directly embeds them into the weights of the FFN module corresponding to the language model layer. This addresses the issues of high computational overhead in mainstream input-space visual prompting methods and insufficient matching of visual and language model layer features in existing memory-space methods, while maintaining parameter efficiency and computational efficiency.

\bibliography{aaai2026}

\clearpage
\section{Appendix}
\subsection{A. Analysis of  DEHVF computational overhead}
In this subsection, we will perform a detailed calculation of the computational cost of DEHVF. Similar to the main text, we assume that the large language model (LLM) has d blocks and a hidden state dimension of h, with an input text length of L and a visual token count of n. Therefore, the number of floating-point operations (FLOPs) for the self-attention module and the feedforward network are $8Lh^2+4L^2h$ and $16Lh^2$, respectively. Since no visual tokens are introduced, the computational cost of the LLM for text token sequence input is $24d(L + n)h^2 + 4d(L + n)^2h$. During training, we embed the fused visual features into the FFN to obtain multimodal information. The additional computations come from three parts: 1) The input embedding network computes compressed features, with input, hidden layer, and output dimensions of $2r$, $2r$, and $r$, respectively, resulting in FLOPs of $12r^3$. 2) The process of computing weights in the hierarchical visual fuser, with k Transformer Blocks and a hidden layer dimension of r, resulting in approximately $3r^3 + 8kr^3$ FLOPs. 3) The additional FLOPs from visual features concatenated into the FFN are $4dLnh$. Therefore, the total FLOPs for DEHVF during training are $24dLh^2 + 4dLn^2h + 12r^3 + 3r^3 + 8kr^3 + 4dLnh$. 

\subsection{B. Experiment Details}
We list the hyperparameters in Table \ref{hyper}. We use task-specific prompt to the input sentence for each downstream task, as shown in Table \ref{prompts}.

\begin{table}[htbp]
\small
\centering
\begin{tabular}{ccccc}
\hline
Names of Hyperparameters     & Value                                                                                                        \\ \hline
Learning Rate                & 2.00E-02                                                                                                     \\
Batch Size                   & 4                                                                                                            \\
Epoch                        & ScienceQA 20, COCO Captions 5                                                                                \\
scaling factor               & 0.01                                                                                                         \\
visual adapter               & \begin{tabular}[c]{@{}c@{}}hidden dimension d = 12, \\ activation GELU\end{tabular}                          \\
Input Embedding Network      & \begin{tabular}[c]{@{}c@{}}input dimension 128, \\ hidden dimension 128, \\ output dimension 64\end{tabular} \\
hierarchical visual fuser    & \begin{tabular}[c]{@{}c@{}}Number of Transformer blocks k=4, \\ hidden dimension d = 64\end{tabular}         \\
length of position embedding & LLaMA-7B 320, LLaMA-13B 420                                                                                  \\ \hline
\end{tabular}
\caption{Hyperparameters on LLaMA.}
\label{hyper}
\end{table}

\begin{table}[htbp]
\small
\centering
\begin{tabular}{ccccc}
\hline
Task          & Input                                                                                                                                                                                             & Output        \\ \hline
ScienceQA     & \begin{tabular}[c]{@{}c@{}}Question: {[}Question{]}\textbackslash{}n \\ Context: {[}Context{]}\textbackslash{}n \\ Options: {[}Choices{]}\textbackslash{}n \\ Reponse: The answer is\end{tabular} & {[}Caption{]} \\
COCO Captions & \begin{tabular}[c]{@{}c@{}}Provide a one-sentence caption\\  for the provided image.\end{tabular}                                                                                                 & {[}Answer{]}  \\ \hline
\end{tabular}
\caption{Input-output formats with task prompts.}
\label{prompts}
\end{table}

\end{document}